\documentclass{KERauth}
\usepackage[round,authoryear]{natbib}
\usepackage{enumerate}
\usepackage{graphicx}

\begin{document}
\KER{1}{24}{00}{0}{2004}{S000000000000000}

\title{One-Class Classification: Taxonomy of Study and Review of Techniques}

\author{SHEHROZ S.KHAN\footnote{Present Address: David R. Cheriton School of Computer Science, University of Waterloo,  Waterloo, Canada N2L 3G1, Tel: +1 519-888-4567 x33123, Fax:  +1 519-885-1208}, MICHAEL G.MADDEN}

\address{College of Engineering and Informatics,
National University of Ireland, Galway,
Republic of Ireland
\\
\email{shehroz@gmail.com, michael.madden@nuigalway.ie}}

\begin{abstract}
One-class classification (OCC) algorithms aim to build classification models when the negative class is either absent, poorly sampled or not well defined. This unique situation constrains the learning of efficient classifiers by defining class boundary just with the knowledge of positive class. The OCC problem has been considered and applied under many research themes, such as outlier/novelty detection and concept learning. In this paper we present a unified view of the general problem of OCC by presenting a taxonomy of study for OCC problems, which is based on the availability of training data, algorithms used and the application domains applied. We further delve into each of the categories of the proposed taxonomy and present a comprehensive literature review of the OCC algorithms, techniques and methodologies with a focus on their significance, limitations and applications. We conclude our paper by discussing some open research problems in the field of OCC and present our vision for future research.
\end{abstract}

\section{Introduction to One-class Classification}
\label{introduction}
The traditional multi-class classification paradigm aims to classify an unknown data object into one of several pre-defined categories (two in the simplest case of binary classification). A problem arises when the unknown data object does not belong to any of those categories. Let us assume that we have a training data set comprising of instances of fruits and vegetables. Any binary classifier can be applied to this problem, if an unknown test object (within the domain of fruits and vegetables e.g. apple or potato) is given for classification. But if the test data object is from an entirely different domain (for example a cat from the category animals), the classifier will always classify the cat as either a fruit or a vegetable, which is a wrong result in both the cases. Sometimes the classification task is just not to allocate a test object into predefined categories but to decide if it belongs to a particular class or not. In the above example an apple belongs to class fruits and the cat does not.

In one-class classification (OCC) \citep{m._j._tax_one-class_2001,m._j._tax_uniform_2001}, one of the classes (which we will arbitrarily refer to as the positive or target class) is well characterized by instances in the training data, while the other class (negative or outlier) has either no instances or very few of them, or they do not form a statistically-representative sample of the negative concept. To motivate the importance of one-class classification, let us consider some scenarios. A situation may occur, for instance, where we want to monitor faults in a machine. A classifier should detect when the machine is showing abnormal/faulty behaviour. Measurements on the normal operation of the machine (positive class training data) are easy to obtain. On the other hand, most possible faults would not have occurred in reality, hence we may have little or no training data for the negative class. Also, we may not want wish to wait until such faults occur as they may involve high cost, machine malfunction or risk to human operators. Another example is the automatic diagnosis of a disease. It is relatively easy to compile positive data (all patients who are known to have a `common' disease) but negative data may be difficult to obtain since other patients in the database cannot be assumed to be negative cases if they have never been tested, and such tests can be expensive. Alternatively, if the disease is `rare' it is difficult to collect positive samples until a sufficiently large group has contracted that disease, which is an unsatisfactory approach. As another example, a traditional binary classifier for text or web pages requires arduous pre-processing to collect negative training examples. For example, in order to construct a \textit{homepage} classifier \citep{yu_pebl:_2002}, sample of homepages (positive training examples) and a sample of non-homepages (negative training examples) need to be gleaned. In this situation, collection of negative training examples is challenging because it may either result in improper sampling of positive and negative classes or it may introduce subjective biases.

The outline of the rest of this paper paper is as follows. Section \ref{occ-mcc} compares OCC with multi-class classification and discusses the performance measures employed for OCC algorithms. Section \ref{related-review} provides an overview of related reviews of OCC. In Section \ref{taxonomy}, we propose a taxonomy for the study of OCC algorithms and present a comprehensive review of the current state of the art and significant research contributions under the branches of the proposed taxonomy. Section \ref{conclusions} concludes our presentation with a discussion on some open research problems and our vision for the future of research in OCC.

\section{One-class Classification Vs Multi-class Classification}
\label{occ-mcc}
In a conventional multi-class classification problem, data from two (or more) classes are available and the decision boundary is supported by the presence of data objects from each class. Most conventional classifiers assume more or less equally balanced data classes and do not work well when any class is severely under-sampled or is completely absent. It appears that \citet{minter_single-class_1975} was the first to use the term `single-class classification' four decades ago, in the context of learning Bayes classifier that requires only labelled data from the ``class of interest''. Much later, \citet{r._moya_one-class_1993} originate the term \textit{One-Class Classification} in their research work. Different researchers have used other terms such as \textit{Outlier Detection\footnote{Readers are advised to refer to detailed literature survey on outlier detection by \citet{chandola_outlier_2009}}} \citep{ritter_outliers_1997}, \textit{Novelty Detection\footnote{Readers are advised to refer to detailed literature survey on novelty detection by \citet{markou_novelty_2003, markou_novelty_2003-1}}} \citep{bishop_novelty_1994},  \textit{Concept Learning} \citep{japkowicz_concept-learning_1999} or \textit{Single Class Classification} \citep{t._munroe_multi-class_2005, yu_single-class_2005, el-yaniv_optimal_2007}. These terms originate as a result of different applications to which one-class classification has been applied.  \citet{juszczak_learning_2006} defines One-Class Classifiers as \textit{class descriptors that are able to learn restricted domains in a multi-dimensional pattern space using primarily just a positive set of examples}.

As observed by \citet{m._j._tax_one-class_2001},  the problems that are encountered in the conventional classification problems, such as the estimation of the classification error, measuring the complexity of a solution, the curse of dimensionality, the generalization of the classification method also appear in OCC and sometimes become even more prominent. As stated earlier, in OCC tasks either the negative data objects are absent or available in limited amount, so only one side of the classification boundary can be determined using only positive data (or some negatives). This makes the problem of one-class classification harder than the problem of conventional two-class classification. The task in OCC is to define a classification boundary around the positive class, such that it accepts as many objects as possible from the positive class, while it minimizes the chance of accepting the outlier objects.  In OCC, since only one side of the boundary can be determined, it is hard to decide on the basis of just one-class how tightly the boundary should fit in each of the directions around the data. It is also harder to decide which features should be used to find the best separation of the positive and outlier class objects. 

\subsection{Measuring Classification Performance of One-class Classifiers}
As mentioned in the work of \citet{m._j._tax_one-class_2001}, a confusion matrix (see Table \ref{table:1}) can be constructed to compute the classification performance of one-class classifiers. To estimate the true error (as is computed for multi-class classifiers), the complete probability density of both the classes should be known. In the case of one-class classification, the probability density of only the positive class is known. This means that only the number of positive class objects which are not accepted by the one-class classifier (i.e. the false negatives, $F^-$) can be minimized. In the absence of examples and sample distribution from outlier class objects, it is not possible to estimate the number of outliers objects that will be accepted by the one-class classifier (the false positives, $F^+$). Furthermore, it can be noted that since $T^+ + F^- = 1 $ and $ T^- + F^+ =1$, the main complication in OCC is that only $T^+$ and $F^-$ can be estimated and nothing is known about $F^+$ and $T^-$. Therefore a limited amount of outlier class data is required to estimate the performance and generalize the classification accuracy of a one-class classifier. However, during testing if the outlier class is not presented in a reasonable proportion, then the actual accuracy values of the one-class classifier could be manipulated and they may not be the true representative of the metric; this point is explored in detail by \citet{g._glavin_analysis_2009}. In such imbalanced dataset scenarios, several other performance metrics can also be  useful, such as f-score, geometric mean etc \citep{nguyen_learning_2009}.

  \begin{table}[!ht]
\begin{center}
  \begin{tabular}{|p{2.75cm}|p{2.5cm}|p{2.5cm}|} \hline
                              & \textbf{Object from target class} & \textbf{Object from outlier class} \\ \hline
\textbf{Classified as a target object}  & True positive, $T^+$    & False positive, $F^+$ \\ \hline
\textbf{Classified as an outlier object}& False negative, $F^-$   & True negative, $T^-$ \\ \hline
  \end{tabular}
  \caption{Confusion Matrix for OCC. Source: \citet{m._j._tax_one-class_2001}.}
  \label{table:1}
\end{center}
  \end{table}

\section{Related Review Work in OCC}
\label{related-review}
In recent years, there has been a considerable amount of research work carried out in the field of OCC. Researchers have proposed several OCC algorithms to deal with various classification problems. \citet{mazhelis_one-class_2006} presents a review of OCC algorithms and analyzed its suitability in the context of mobile-masquerader detection. In that paper, Mazhelis proposes a taxonomy of one-class classifiers classification techniques based on:
\begin{enumerate}[(i)]
 \item The internal model used by classifier (density, reconstruction or boundary based)
 \item The type of data (numeric or symbolic), and
 \item The ability of classifiers to take into account temporal relations among feature (yes or no). 
 \end{enumerate}
 
Mazhelis's survey of OCC describes a lot of algorithms and techniques; however it covers a sub-spectrum of the problems in the field of OCC. As we describe in subsequent sections, one-class classification techniques have been developed and used by various researchers by different names in different contexts. The survey presented by \citet{mazhelis_one-class_2006} proposes a taxonomy suitable to evaluate the applicability of OCC to the specific application domain of mobile-masquerader detection. 

\citet{brew_evaluation_2007} present a review of several OCC algorithms along with Gaussian Mixture models for the Speaker Verification problem. Their main work revolves around front-end processing and feature extraction from speech data, and speaker and imposter modelling and using them in OCC framework. \citet{kennedy_credit_2009} discuss some issues related to OCC and presents a review of several OCC approaches that includes statistical, neural networks and support vector machines based methods. They also discuss the importance of including non-target data for building OCC models and use this study for developing credit-scoring system to identify good and bad creditors. \citet{bergamini_combining_2009} present a brief overview of OCC algorithms for the biometric applications. They identified two broad categories for OCC development as density approaches and boundary approaches. \citet{m._bartkowiak_anomaly_2011} presents survey of research on Anomaly, Outlier and OCC. The research survey is mostly focussed on research and their applications of OCC for detecting unknown behaviour. \citet{s._khan_survey_2009} present a short survey of the recent trends in the field of OCC, wherein they present a taxonomy for the study of OCC methods. Their taxonomy is based on availability of training data, methodology used and application domains applied.

This publication is an extension of the work of \citet{s._khan_survey_2009}, and is more comprehensive, in-depth and detailed. The survey in this publication identifies some important research areas, raises several open questions in the study of OCC and discusses significant contributions made by researchers (see Section \ref{taxonomy}). In this work, we neither restrict the review of literature pertaining to OCC to a particular application domain, nor to specific algorithms that are dependent on type of the data or model. Our aim is to cover as many algorithms, designs, contexts and applications where OCC has been applied in multiple ways (as shown by examples in Section \ref{introduction}). This publication does not intend to duplicate or re-state previous review work; little of the research work presented here may be found in the past surveys of \citet{mazhelis_one-class_2006}, \citet{brew_evaluation_2007}, \citet{kennedy_credit_2009}, \citet{bergamini_combining_2009} or  \citet{m._bartkowiak_anomaly_2011}. Moreover, this publication encompasses a broader definition of OCC than many. 

\section{Proposed Taxonomy}
\label{taxonomy}
\begin{figure}
  \centering
    \includegraphics[width=5in, height=4in]{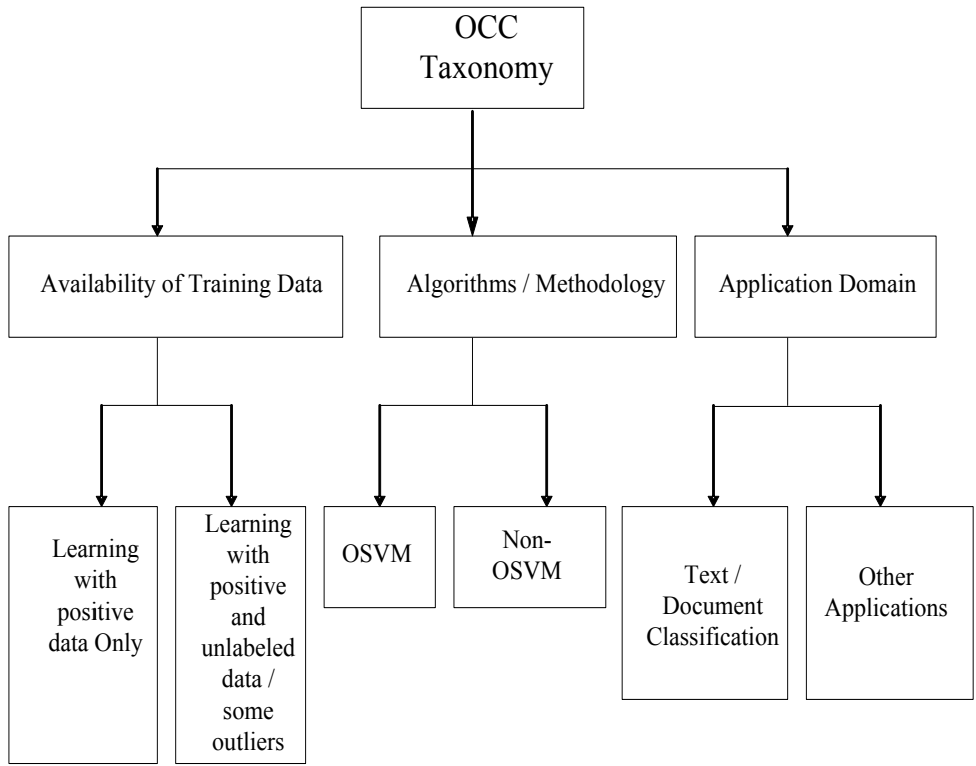}
  \caption{Our Taxonomy for the Study of OCC Techniques.}
  \label{fig:1}
\end{figure}

Based on the research work carried out in the field of OCC using different algorithms, methodologies and application domains, we present a unified approach to OCC by proposing a taxonomy for the study of OCC problems. The taxonomy is divided into three broad categories (see Figure \ref{fig:1}):
 
\begin{enumerate}[(i)]
 \item \textit{Availability of Training Data}: Learning with positive data only or learning with positive and unlabeled data and/or some amount of outlier samples.
 \item \textit{Methodology Used}: Algorithms based on One-class Support Vector Machines (OSVMs) or methodologies based on algorithms other than OSVMs.
 \item \textit{Application Domain}: OCC applied in the field of text/document classification or in other application domains.
\end{enumerate}

The proposed categories are not mutually exclusive, so there may be some overlapping among the research carried out in each of these categories. However, they cover almost all of the major research conducted by using the concept of OCC in various contexts and application domains. The key contributions in most OCC research fall into one of the above-mentioned categories. In the subsequent subsections, we will consider each of these categories in detail.

\subsection{Category 1: Availability of Training Data}
Th availability of training data plays a pivotal role in any OCC algorithm. Researchers have studied OCC extensively under three broad categories: 
\begin{enumerate}[(a)]
 \item Learning with positive examples only. 
 \item Learning with positive examples and some amount of poorly sampled negative examples or artificially generated outliers.
 \item Learning with positive and unlabeled data. 
\end{enumerate}

Category (c) has been a matter of much research interest among the text/document classification community \citep{liu_building_2003,li_learning_2003,lee_learning_2003} that will be discussed in detail in Section \ref{text}.  

\citet{m._j._tax_data_1999, m._j._tax_support_1999} and \citet{scholkopf_estimating_1999} have developed various algorithms based on support vector machines to tackle the problem of OCC using positive examples only; for a detailed discussion on them, refer to Section \ref{osvm}. The main idea behind these strategies is to construct a decision boundary around the positive data so as to differentiate them from the outlier/negative data.

For many learning tasks, labelled examples are rare while numerous unlabeled examples are easily available. The problem of learning with the help of unlabeled data given a small set of labelled examples was studied by \citet{blum_combining_1998} by using the concept of co-training. The co-training approach can be applied when a data set has natural separation of their features and classifiers are built incrementally on them. Blum and Mitchell demonstrate the use of co-training methods to train the classifiers in the application of text classification.  Under the assumptions that each set of the features is sufficient for classification, and the feature sets of each instance are conditionally independent given the class, they provide PAC (Probably Approximately Correct) learning \citep{g._valiant_theory_1984} guarantees on learning from labelled and unlabeled data and prove that unlabeled examples can boost accuracy. \citet{denis_pac_1998} was the first to conduct a theoretical study of PAC learning from positive and unlabeled data. Denis proved that many concept classes, specifically those that are learnable from statistical queries, can be efficiently learned in a PAC framework using positive and unlabeled data. However, the trade-off is a considerable increase in the number of examples needed to achieve learning, although it remains polynomial in size. \citet{de_comite_positive_1999} give evidence with both theoretical and empirical arguments that positive examples and unlabeled examples can boost accuracy of many machine learning algorithms. They noted that the learning with positive and unlabeled data is possible when the weight of the target concept (i.e. the ratio of positive examples) is known by the learner, which in turn can be estimated from a small set of labelled examples. \citet{muggleton_learning_2001} presents a theoretical study in the Bayesian framework where the distribution of functions and examples are assumed to be known. \citet{liu_partially_2002} extend Muggleton's result to the noisy case; they present sample complexity results for learning by maximizing the number of unlabeled examples labelled as negative while constraining the classifier to label all the positive examples correctly. Further details on the research carried out on training classifiers with labelled positive and unlabeled data is presented in Section \ref{text}.

\subsection{Category 2: Algorithms Used}
Most of the major OCC algorithms development can be classified under two broad categories, as has been done either using:
\begin{itemize}
  \item  One-class Support Vector Machines (OSVMs), or
  \item  Non-OSVMs methods (including various flavours of neural networks, decision trees, nearest neighbours and others). 
\end{itemize}

It may appear at first that this category presents a biased view, by classifying algorithms according to whether or not they are based on OSVM vs Non-OSVM. However we have found that the advancements, applications, significance and difference that OSVM based algorithms have shown opens it up as a separate research area in its own right. Nonetheless, Non-OSVM based OCC algorithms have also been used for tackling specific research problems; the details are presented in the following sub-sections.

\subsubsection{One-class Support Vector Machine (OSVM)}
\label{osvm}
\begin{figure}
  \centering
    \includegraphics[width=2.5in]{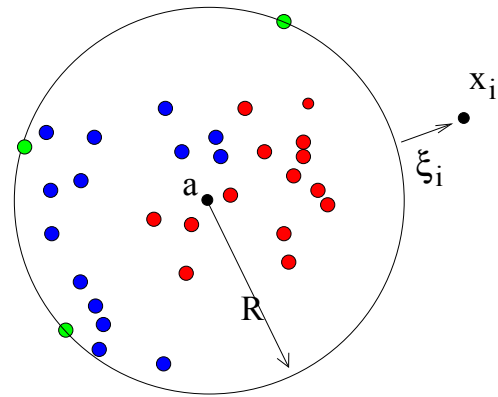}
  \caption{The hyper-sphere containing the target data, with centre a and radius R. Three objects are on the boundary are the support vectors. One object $x_i$ is outlier and has $\xi > 0$. Source: \citet{m._j._tax_one-class_2001}.}
  \label{fig:2}
\end{figure}

\begin{figure}
  \centering
    \includegraphics[width=3.5in]{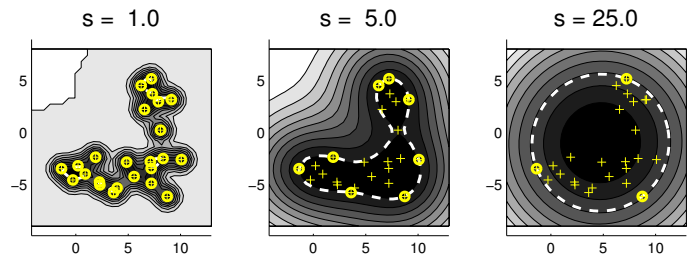}
  \caption{Data description trained on a banana-shaped data set. The kernel is a Gaussian kernel with different width sizes $s$. Support vectors are indicated by the solid circles; the dashed line is the description boundary. Source: \citet{m._j._tax_one-class_2001}.}
  \label{fig:3}
\end{figure}

\citet{m._j._tax_data_1999, m._j._tax_support_1999} seek to solve the problem of OCC by distinguishing the positive class from all other possible data objects in the pattern space. They constructed a hyper-sphere around the positive class data that encompasses almost all points in the data set with the minimum radius. This method is called the Support Vector Data Description (SVDD) (See Figure \ref{fig:2}). The SVDD classifier rejects a given test point as outlier if it falls outside the hyper- sphere. However, SVDD can reject some fraction of positively labelled data when the volume of the hyper-sphere decreases. The hyper-sphere model of the SVDD can be made more flexible by introducing kernel functions. \citet{m._j._tax_one-class_2001} considers Polynomial and a Gaussian kernel and found that the Gaussian kernel works better for most datasets considered (Figure \ref{fig:3}). Tax uses different values for the width of the kernel. The larger the width of the kernel, the fewer support vectors are selected and the description becomes more spherical. Also, using the Gaussian kernel instead of the Polynomial kernel results in tighter descriptions, but it requires more data to support more flexible boundary. Tax's method becomes inefficient when the data set has high dimension. It also does not work well when large variations in density exist among the positive-class objects; in such case, it starts rejecting the low-density target points as outliers.  \citet{m._j._tax_one-class_2001} demonstrates the usefulness of the approach on machine fault diagnostic data and handwritten digit data.

\citet{m._j._tax_uniform_2001} propose a sophisticated method which uses artificially generated outliers to optimize the OSVM parameters in order to balance between over-fitting and under-fitting. The fraction of the outliers accepted by the classifier is an estimate of the volume of the feature space covered by the classifier. To compute the error without the use of outlier examples, they uniformly generate artificial outliers in and around the target class. If a hyper-cube is used, then in high dimensional feature space it becomes infeasible.  In that case, the outlier objects generated from a hyper-cube will have very low probability to be accepted by the classifier. The volume in which the artificial outliers are generated has to fit as tightly as possible around the target class. To make this procedure applicable in high dimensional feature spaces, Tax and Duin propose to generate outliers uniformly in a hyper-sphere. This is done by transforming objects generated from a Gaussian distribution. Their experiments suggest that the procedure to artificially generate outliers in a hyper-sphere is feasible for up to 30 dimensions.
 
\citet{scholkopf_support_2000, scholkopf_sv_1999} present an alternative approach to SVDD. In their method they construct a hyper-plane instead of a hyper-sphere around the data, such that this hyper-plane is maximally distant from the origin and can separate the regions that contain no data. They propose to use a binary function that returns +1 in `small' region containing the data and -1 elsewhere. They introduce a variable that controls the effect of outliers i.e. the hardness or softness of the boundary around the data. \citet{scholkopf_sv_1999} suggest the use of different kernels, corresponding to a variety of non-linear estimators. In practical implementations, the method of  Sch\"{o}lkopf et al. and the SVDD method of \citet{m._j._tax_uniform_2001} operate comparably and both perform best when the Gaussian kernel is used. As mentioned by Campbell and Bennett \citet{campbell_linear_2001}, the origin plays a crucial role in the methods of both Sch\"{o}lkopf et al. and Tax \& Duin, which is a drawback since the origin effectively acts as a prior for where the abnormal class instances are assumed to lie; this is termed the problem of origin. \citet{scholkopf_sv_1999} have tested their method on both synthetic and real-world data, including the US Postal Services dataset of handwritten digits. Their experiments show that the algorithm indeed extracts data objects that are difficult to be assigned to their respective classes and a number of outliers were in fact identified. 

\begin{figure}[ht]
\centering
  \includegraphics[width=3.5in]{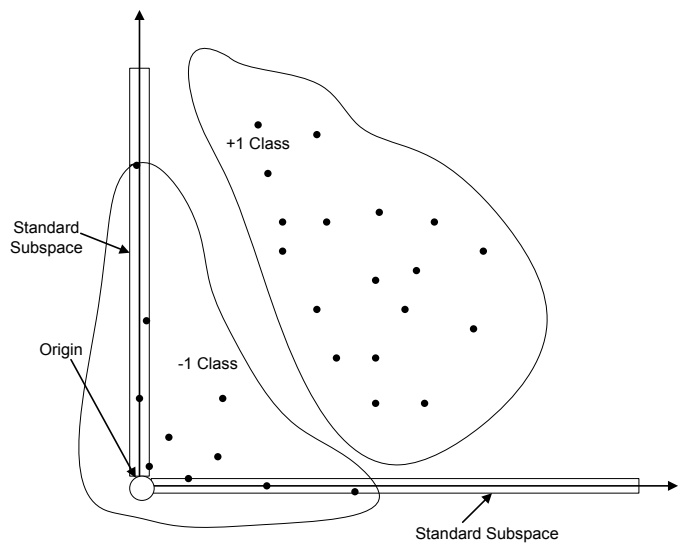}
  \caption{Outlier SVM Classifier. The origin and small subspaces are the original members of the second class. Source: \citet{m._manevitz_one-class_2001}.}
  \label{fig:4a}
\end{figure}

\begin{figure}[ht]
\centering
  \includegraphics[width=3.5in]{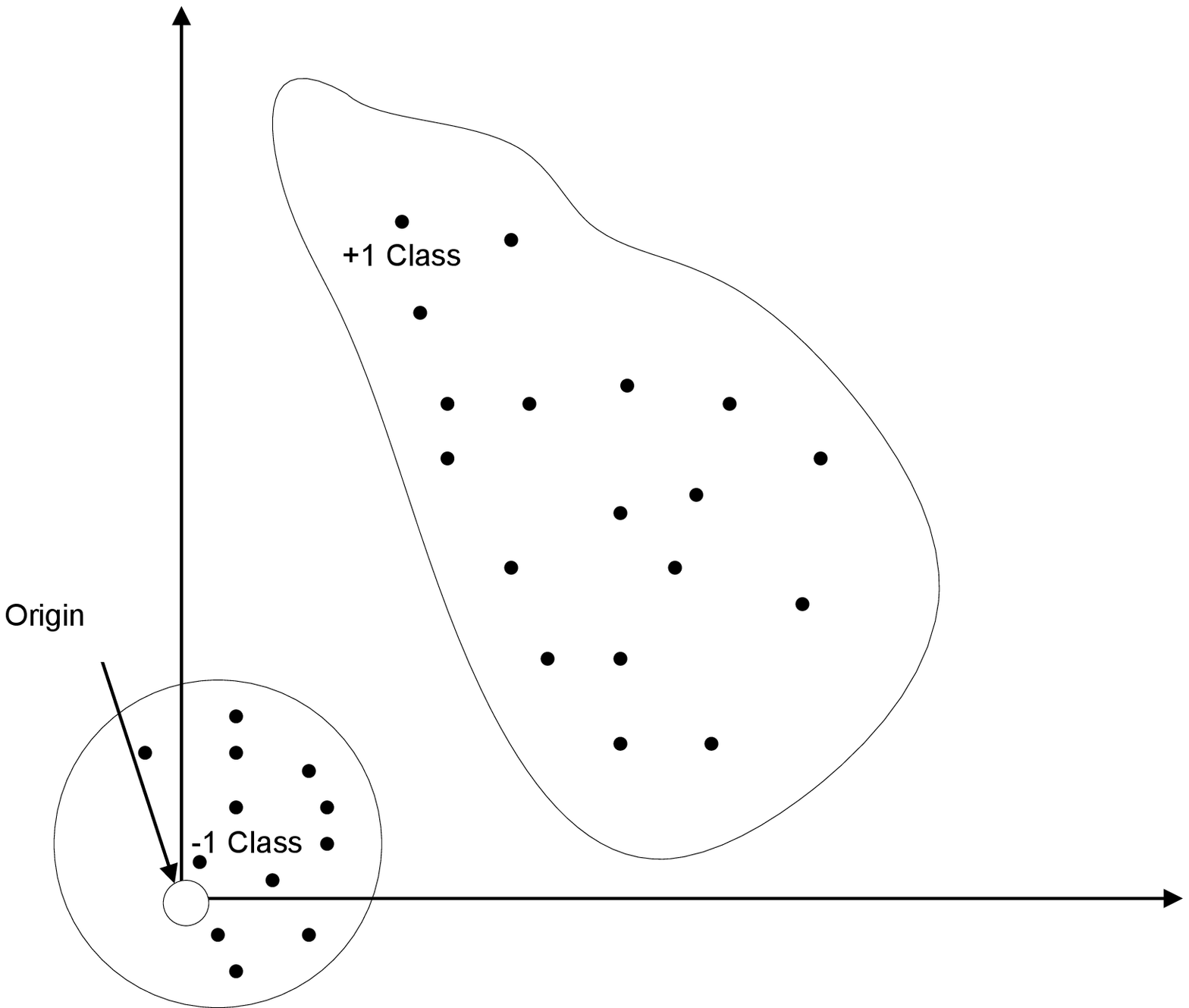}
  \caption{Improved OSVM. Source: \citet{li_improving_2003}.}
  \label{fig:4b}
\end{figure}

\citet{m._manevitz_one-class_2001} investigate the use of one-class SVM for information retrieval. Their paper proposes a different version of the one-class SVM than that proposed by \citet{scholkopf_estimating_1999}; the method of Manevitz \& Yousef is based on identifying outlier data that is representative of the second class. The idea of their methodology is to work first in the feature space, and assume that not only the origin is member of the outlier class, but also all  data points close to the origin are considered as noise or outliers (see Figure \ref{fig:4a}). Geometrically speaking, the vectors lying on standard sub-spaces of small dimension i.e. axes, faces, etc., are to be treated as outliers. Hence, if a vector has few non-zero entries, then this indicates that the data object shares very few items with the chosen feature subset of the database and will be treated as an outlier. Linear, sigmoid, polynomial and radial basis kernels were used in their work. \citet{m._manevitz_one-class_2001} evaluate the results on the Reuters data set\footnote{http://www.daviddlewis.com/resources/testcollections/reuters21578/ [Accessed: Jan-2012]}  using the 10 most frequent categories.  Their results are generally somewhat worse than the OSVM \citep{scholkopf_sv_1999}. However they observe that when the number of categories are increased, their version of OSVM obtains better results. \citet{li_improving_2003} present an improved version of the approach of \citet{scholkopf_sv_1999} for detecting anomaly in an intrusion detection system, with higher accuracy. Their idea is to consider all points ``close enough'' to the origin as outliers and not just the origin as the member of second class (see Figure \ref{fig:4b}). Zhao et al. \citep{zhao_customer_2005} use this method for customer churn prediction for the wireless industry data. They investigate the performance of different kernel functions for this version of one-class SVM, and show that the Gaussian kernel function can detect more churners than the Polynomial and Linear kernel. 

An extension to the work of \citet{m._j._tax_data_1999, m._j._tax_support_1999} and \citet{scholkopf_support_2000} is proposed by  \citet{campbell_linear_2001}. They present a kernel OCC algorithm that uses linear programming techniques instead of quadratic programming. They construct a surface in the input space that envelopes the data, such that the data points within this surface are considered targets and outside it are regarded as outliers. In the feature space, this problem condenses to finding a hyper-plane which is pulled onto the data points and the margin remains either positive or zero. To fit the hyper-plane as tightly as possible, the mean value of the output of the function is minimized. To accommodate outliers, a soft margin is introduced around the hyper-plane. Their algorithm avoids the problem of the origin (stated earlier) by attracting the hyper-plane towards the centre of data distribution rather than by repelling it away from a point outside the data distribution. In their work, different kernels are used to create hyper-planes and they show that the Radial Basis Function kernel can produce closed boundaries in input space while other kernels may not \citep{campbell_linear_2001}. A drawback of their method is that it is highly dependent on the choice of kernel width parameter, $\sigma$. However, if the data size is large and contains some outliers then $\sigma$ can be estimated. They show their results on artificial data set, Biomedical Data and Condition Monitoring data for machine fault diagnosis. 

\citet{yang_one-class_2007} apply particle swarm optimization to calibrate the parameters of OSVM \citep{scholkopf_support_2000} and find experimentally that their method either matches or surpass the performance of OSVM with parameters optimized using grid search method, while using lower CPU time.  \citet{tian_anomaly_2010} propose a refinement to the Sch\"{o}lkopf's OSVM model \citep{scholkopf_support_2000} by searching optimal parameters using particle swarm optimization algorithm \citep{kennedy_particle_1995} and improving the original decision function with a boundary movement. Their experiments show that after adjusting the threshold, the final decision function gives a higher detection rate and a lower rejection rate.

\citet{luo_research_2007} extends the work of \citet{m._j._tax_uniform_2001} to propose a cost-sensitive OSVM algorithm called Frequency-Based SVDD (F-SVDD) and Write-Related SVDD (WS-SVDD) for intrusion detection problem. The SVDD method gives equal cost to classification errors, whereas F-SVDD gives higher cost to frequent short sequences occurring during system calls and WS-SVDD gives different costs to different system calls. Their experiments suggest that giving different cost or importance to system users than to processes results in higher performance in intrusion detection than SVDD. \citet{yang_brain_2010} propose a neighbourhood-based OSVM method for fMRI (functional magnetic resonance imaging) data, where the objective is to classify individauls as having schizophrenia or not, based on fMRI scans of their brains. In their formulation of OSVM, they assume the neighbourhood consistency hypothesis used by \citet{f._chen_new_2002}. By integrating it with OSVM, they compute primal values which denote distance between points and hyper-plane in kernel space. For each voxel in an fMRI image, a new decision value is computed using the primal values and their neighbours. If this decision value is greater than a given threshold then it is regarded as activated voxel, otherwise it is regarded as non-activated. Their experiments on various brain fMRI data sets show that it gives more stable results than K-Means and fuzzy K-Means clustering algorithms.

\begin{figure}
  \centering
    \includegraphics[width=4in]{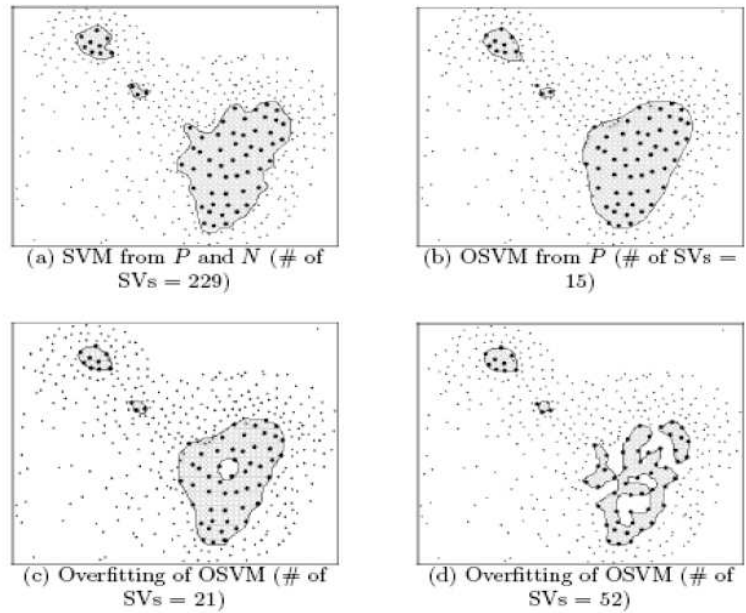}
  \caption{Boundaries of SVM and OSVM on a synthetic data set: big dots: positive data, small dots: negative data. Source: \citet{yu_single-class_2005}.}
  \label{fig:5}
  \end{figure}

\citet{yu_single-class_2005} proposes a one-class classification algorithm with SVMs using positive and unlabeled data and without labelled negative data and discuss some of the limitations of other OSVM-based OCC algorithms \citep{m._j._tax_uniform_2001, m._manevitz_one-class_2001}. In assessing the performance of OSVMs under a scenario of learning with unlabeled data and no negative examples, Yu comments that to induce an accurate class boundary around the positive data set, OSVM requires a larger amount of training data. The support vectors in such a case come only from positive examples and cannot create a proper class boundary, which also leads to either over-fitting or under-fitting of the data (See Figure \ref{fig:5}). Yu notes that when the numbers of support vectors in OSVM were increased, it overfits the data rather than being more accurate. \citet{yu_svmc:_2003} presents an OCC algorithm called Mapping Convergence (MC) to induce accurate class boundary around the positive data set in the presence of unlabeled data and without negative examples. The algorithm has two phases: mapping and convergence. In the first phase, a weak classifier (e.g. \citet{rocchio_relevant_1971}) is used to extract strong negatives (those that are far from the class boundary of the positive data) from the unlabeled data. In the second phase, a base classifier (e.g. SVM) is used iteratively to maximize the margin between positive and strong negatives for better approximation of the class boundary. \citet{yu_svmc:_2003} also presents another algorithm called Support Vector Mapping Convergence (SVMC) that works faster than the MC algorithm. At every iteration, SVMC only uses minimal data so that the accuracy of class boundary is not degraded and the training time of SVM is also saved. However, the final class boundary is slightly less accurate than the one obtained by employing MC. They show that MC and SVMC perform better than other OSVM algorithm and can generate accurate boundaries comparable to standard SVM with fully labelled data.

\citet{y._hao_fuzzy_2008} incorporates the concept of fuzzy set theory into OSVM \citep{scholkopf_sv_1999} in order to deal with the problem that standard OSVM can be sensitive to outliers and noise. The main idea of Hao's method is that different training data objects may contribute differently to the classification, which can be estimated using fuzzy membership. Two versions of the fuzzy one-class classifier are proposed: (i) a crisp hyper-plane is constructed to separate target class from origin and fuzzy membership is associated to training data, where noisy and outlier data can be given low fuzzy membership values, (ii) a fuzzy hyper-plane is constructed to discriminate target class from others. Ho uses Gaussian kernels with different parameter settings and test their method on handwritten digits problem. Another fuzzy OSVM classifier is proposed by \citet{s._choi_video_2004} for generating visually salient and semantically important video segments. Their algorithm gives different weights to the importance measure of the video segments and then estimates their support. They demonstrate the performance of their algorithm on several synthesized datasets and different types of videos.

\subsubsection{One-Class Classifiers other than OSVMs}
In this section we review some major OCC algorithms that are based on one-class ensembles, neural networks, decision trees, nearest neighbours, Bayesian classifiers and other methods.

\paragraph{One-Class Classifier Ensemble.}
As in traditional multi-class classification problems, one one-class classifier might not capture all characteristics of the data. However, using just the best classifier and discarding the classifiers with poorer performance might waste valuable information \citep{wolpert_stacked_1992}. To improve the performance of different classifiers which may differ in complexity or in the underlying training algorithm used to construct them, an ensemble of classifiers is a viable solution. This may serve to increase the performance and also the robustness of the classification \citep{j._c._sharkey_how_1995}. Classifiers are commonly ensembled to provide a combined decision by averaging the estimated posterior probabilities. This simple algorithm is known to give good results for multi-class problems \citep{tanigushi_averaging_1997}. In the case of one-class classifiers, the situation is different. One-class classifiers cannot directly provide posterior probabilities for target (positive class) objects, because accurate information on the distribution of the outlier data is not available, as has been discussed earlier in this paper. In most cases, by assuming that the outliers are uniformly distributed, the posterior probability can be estimated. \citet{m._j._tax_one-class_2001} mentions that in some OCC methods, distance is estimated instead of probability, and if there exists a combination of distance and probability outputs, they should be standardized before they can be combined. Having done so, the same types of combining rules as in conventional classification ensembles can be used.  \citet{m._j._tax_combining_2001} investigate the influence of the feature sets, their inter-dependence and the type of one-class classifiers for the best choice of  combination rules. They use a Normal density and a mixture of Gaussians and the Parzen density estimation \citep{bishop_novelty_1994} as two types of one-class classifiers. They use four models, the SVDD \citep{m._j._tax_support_1999}, K-means clustering, K-center method \citep{ypma_support_1998} and an auto-encoder neural network \citep{japkowicz_concept-learning_1999}. In their experiments, the Parzen density estimator emerges as the best individual one-class classifier on the handwritten digit pixel dataset \footnote{ftp://ftp.ics.uci.edu/pub/machine-learning-databases/mfeat/ [Accessed August 2013]}. \citet{m._j._tax_one-class_2001} show that combining classifiers trained on different feature spaces is useful. In their experiments, the product combination rule gives the best results while the mean combination rule suffers from the fact that the area covered by the target set tends to be overestimated.
 
\citet{lai_combining_2002} study combining one-class classifier for image database retrieval and show that combining SVDD-based classifiers improve the retrieval precision. \citet{juszczak_combining_2004} extend combining one-class classifierd for classifying missing data. Their idea is to form an ensemble of one-class classifiers trained on each feature or each pre-selected group of features, or to compute a dissimilarity representation from features. The ensemble is able to predict missing feature values based on the remaining classifiers. As compared to standard methods, their method is more flexible, since it requires significantly fewer classifiers and does not require re-training of the system whenever missing feature values occur.  \citet{juszczak_combining_2004} also show that their method is robust to small sample size problems due to splitting the classification problem into several smaller ones. They compare the performance of their proposed ensemble method with standard methods used with missing features values problem on several UCI datasets \citep{Bache+Lichman:2013}.

\citet{ban_implementing_2006} address the problem of building a multi-class classifier based on an ensemble of one-class classifiers by studying two kinds of one-class classifiers, namely, SVDD \citep{m._j._tax_support_1999} and Kernel Principal Component Analysis \citep{scholkopf_nonlinear_1998}. They construct a minimum-distance-based classifier from an ensemble of one-class classifiers that is trained from each class and assigns a test data object to a given class based on its prototype distance. Their method gives comparable performance to that of SVMs on some benchmark datasets; however it is heavily dependent on the algorithm parameters. They also comment that their process could lead to faster training and better generalization performance provided appropriate parameters are chosen. 

\citet{pekalska_combining_2004} use the proximity of target object to its class as a `dissimilarity representation' (DR) and show that the discriminative properties of various DRs can be enhanced by combining them properly. They use three types of one-class classifier, namely Nearest Neighbour, Generalized Mean Class and Linear Programming Dissimilarity Data Description. They make two types of ensembles: (i) combine different DR from individual one-class classifiers into one representation after proper scaling using fixed rules, for e.g. average, product and train single one-class classifier based on this information, (ii) combine different DR of training objects over several base classifiers using majority voting rule. Their results show that both methods perform significantly better than the OCC trained with a single representation. \citet{nanni_experimental_2006} studies combining several one-class classifiers using the random subspace method \citep{k._ho_random_1998} for the problem of online signature verification. Nanni's method generates new training sets by selecting features randomly and then employing one-class classification on them; the final results of these classifiers are combined through the max rule. Nanni uses several one-class classifiers: Gaussian model description; Mixture of Gaussian Descriptions; Nearest Neighbour Method Description; PCA Description (PCAD); Linear Programming Description (LPD); SVDD; and Parzen Window Classifier. It is shown that fusion of various classifiers can reduce the error and the best fusion method is the combination of LPD and PCAD.  \citet{cheplygina_pruned_2011} propose to apply pruning to random sub-spaces of one-class classifiers. Their results show that pruned ensembles give better and more stable performance than the complete ensemble.  \citet{bergamini_fusion_2008} present the use of OSVM for biometric fusion where very low false acceptance rates are required. They suggest a fusion system that may be employed at the level of feature selection, score-matching or decision-making. A normalization step is used before combining scores from different classifiers. Bergamini et al. use z-score normalization, min-max normalization, and column norm normalization, and classifiers are combined using various ensemble rules. They find that min-max normalization with a weighted sum gives the best results on NIST Biometric Scores Set. \citet{d._gesu_combining_2007} present an ensemble method of combining one-class fuzzy KNN classifiers. Their classifier-combining method is based on a genetic algorithm optimization procedure by using different similarity measures. Ges\`{u} and Bosco test their method on two categorical datasets and show that whenever the optimal parameters are found, fuzzy combination of one-class classifiers may improve the overall recognition rate. 

Bagging \citep{breiman_bagging_1996} is an ensemble method (for multi-class classification problems) that combines multiple classifiers on re-sampled data to improve classification accuracy.  \citet{li_integrating_2006} extends the one-class information bottleneck method for information retrieval by introducing bagging ensemble learning. The proposed ensemble emphasizes different parts of the data and results from different parameter settings are aggregated to give a final ranking, and the experimental results show improvements in image retrieval applications.  \citet{d._shieh_ensembles_2009} propose an ensemble method for combining OSVM \citep{m._j._tax_data_1999} using bagging. However, bagging is useful when the classifiers are unstable and small changes in the training data can cause large changes in the classifier outputs \citep{bauer_empirical_1999}. OSVM is not an unstable classifier as its estimated boundary always encloses the positive class, therefore directly applying bagging on OSVM is not useful.  \citet{d._shieh_ensembles_2009} propose a kernel density estimation method to give weights to the training data objects, such that the outliers get the least weights and the positive class members get higher weights for creating bootstrap samples. Their experiments on synthetic and real datasets show that bagging OSVMs achieve higher true positive rate. They also mention that such a method is useful in application where low false positive rates are required, such as disease diagnosis. 

Boosting methods are widely used in traditional classification problems \citep{g._dietterich_experimental_2000} for their high accuracy and ease of implementation.  \citet{ratsch_constructing_2002} propose a boosting-like one-class classification algorithm based on a technique called barrier optimization \citep{g._luenberger_linear_1984}.  They also show, through an equivalence of mathematical programs, that a support vector algorithm can be translated into an equivalent boosting-like algorithm and vice versa. It has been pointed out by \citet{e._schapire_boosting_1998} that boosting and SVMs are `essentially the same' except for the way they measure the margin or the way they optimize their weight vector: SVMs use the $l_2$-norm to implicitly compute scalar products in feature space with the help of kernel trick, whereas boosting employs the $l_1$-norm to perform computation explicitly in the feature space. Schapire et al. comment that SVMs can be thought of as a `boosting' approach in high dimensional feature space spanned by the base hypotheses. \citet{ratsch_constructing_2002} exemplify this translation procedure for a new algorithm called the one-class leveraging. Building on barrier methods, a function is returned which is a convex combination of the base hypotheses that leads to the detection of outliers. They comment that the prior knowledge that is used by boosting algorithms for the choice of weak learners can be used in one-class classification and show the usefulness of their results on artificially generated toy data and the US Postal Service database of handwritten characters. 

\paragraph{Neural Networks.}
\citet{de_ridder_experimental_1998} conduct an experimental comparison of various OCC algorithms. They compare a number of unsupervised methods from classical pattern recognition to several variations of a standard shared weight supervised neural network \citep{l._cun_backpropagation_1989} and show that adding a hidden layer with radial basis function improves performance.  \citet{m._manevitz_document_2000} show that a simple neural network can be trained to filter documents when only positive information is available. They design a \textit{bottleneck} filter that uses a basic feed-forward neural network that can incorporate the restriction of availability of only positive examples. They chose three level network with $m$ input neurons, $m$ output neurons and $k$ hidden neurons, where $k < m$. The network is trained using a standard back-propagation algorithm \citep{e._rumelhart_parallel_1986} to learn the identity function on the positive examples. The idea is that while the bottleneck prevents learning the full identity function on $m$-space, the identity on the small set of examples is in fact learnable. The set of vectors for which the network acts as the identity function is more like a sub-space which is similar to the trained set. For testing a given vector, it is shown to the network and if the result is the identity, the vector is deemed interesting (i.e. positive class) otherwise it is deemed an outlier.  \citet{m._manevitz_learning_2000} apply the auto-associator neural network to document classification problem. During training, they check the performance values of the test set at different levels of error. The training process is stopped at the point where the performance starts a steep decline. A secondary analysis is then performed to determine an optimal threshold. Manevitz and Yousef test the method and compare it with a number of competing approaches (i.e. Neural Network; Na\"{i}ve Bayes (NB); Nearest Neighbour; Prototype algorithm) and conclude that it outperforms them.  

\citet{skabar_single-class_2003} describes how to learn a classifier based on feed-forward neural network using positive examples and corpus of unlabeled data containing both positive and negative examples. In a conventional feed-forward binary neural network classifier, positive examples are labelled as 1 and negative examples as 0. The output of the network represents the probability that an unknown example belongs to the target class, with a threshold of 0.5 typically used to decide which class an unknown sample belongs to. However, in this case, since unlabeled data can contain some unlabeled positive examples, the output of the trained neural network may be less than or equal to the actual probability that an example belongs to the positive class. If it is assumed that the labelled positive examples adequately represent the positive concept, it can be hypothesized that the neural network will be able to draw a class boundary between negative and positive examples. Skabar shows the application of the technique to the prediction of mineral deposit location. 

\paragraph{Decision Trees.}
Several researchers have used decision tress to classify positive samples from a corpus of unlabeled examples.  \citet{de_comite_positive_1999} present experimental results showing that positive examples and unlabeled data can efficiently boost accuracy of the statistical query learning algorithms for monotone conjunctions in the presence of classification noise and present experimental results for decision tree induction. They modify standard C4.5 algorithm \citep{r._quinlan_c4.5:_1993} to get an algorithm that uses unlabeled and positive data and show the relevance of their method on UCI datasets.  \citet{letouzey_learning_2000} design an algorithm which is based on positive statistical queries (estimates for probabilities over the set of positive instances) and instance statistical queries (estimates for probabilities over the instance space). The algorithm guesses the weight of the target concept i.e. the ratio of positive instances in the instance space and then uses a hypothesis testing algorithm. They show that the algorithm can be estimated in polynomial time and is learnable from positive statistical queries and instance statistical queries only. Then, they design a decision tree induction algorithm, called POSC4.5, using only positive and unlabeled data and present experimental results on UCI datasets that are comparable to the C4.5 algorithm. \citet{yu_single-class_2005} comments that such rule learning methods are simple and efficient for learning nominal features but are tricky to use for problems of continuous features, high dimensions, or sparse instance spaces.  \citet{li_bagging_2008} perform bagging ensemble on POSC4.5 and classify test samples using the majority voting rule.  Their result on UCI datasets shows that the classification accuracy and robustness of POSC4.5 could be improved by applying their technique. Based on very fast decision trees (VFDT) \citep{domingos_mining_2000} and POSC4.5,  \citet{li_ocvfdt:_2009} propose a one-class VFDT for data streams with applications to credit fraud detection and intrusion detection. They state that by using the proposed algorithm, even if 80\% of the data is unlabeled, the performance of one-class VFDT is very close to the standard VFDT algorithm.  \citet{desir_random_2012} propose a one-class Random Forest algorithm that internally conjoins bagging and random feature selection (RFS) for decision trees. They note that the number of artificial outliers to be generated to convert an one-class classifier to a binary classifier can be exponential with respect to the size of the feature space and availability of positive data objects. They propose a novel algorithm to generate outliers in small feature spaces by combining RFS and Random Subspace methods. Their method aims at generating more artificial outliers in the regions where the target data objects are sparsely populated and less in areas where the density of target data objects is high. They compare their method against OSVM, Gaussian Estimator, Parzen Windows and Mixture of Gaussian Models on an image medical dataset and two UCI datasets and show that it performs equally well or better than the other algorithms.

\begin{figure}
  \centering
    \includegraphics[width=3.25in]{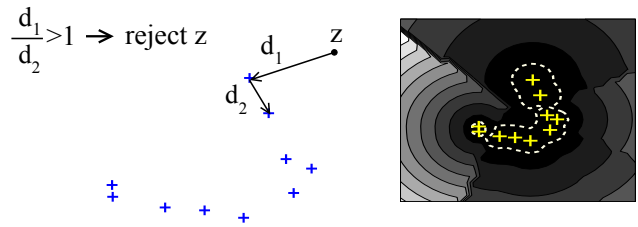}
  \caption{The Nearest Neighbour Data Description. Source: \citet{m._j._tax_one-class_2001}.}
  \label{fig:6}
\end{figure}

\paragraph{Nearest Neighbours.}
\citet{m._j._tax_one-class_2001} presents a one-class Nearest Neighbour method, called Nearest Neighbour Description ($NN$-$d$), where a test object $z$ is accepted as a member of target class provided that its local density is greater than or equal to the local density of its nearest neighbour in the training set. The first nearest neighbour is used for the local density estimation ($1$-$NN$). The following acceptance function is used:
\begin{equation}
  f_{NN^{tr}}(z)=I(\frac{\|z-NN^{tr}(z)\|}{\|NN^{tr}(z)-NN^{tr}(NN^{tr}(z))\|})
\end{equation}
  
which shows that the distance from object $z$ to its nearest neighbour in the training set $NN^{tr}(z)$  is compared to the distance from this nearest neighbour $NN^{tr}(z)$ to its nearest neighbour (see Figure \ref{fig:6}).

The $NN$-$d$ has several predefined choices to tune various parameters. Different numbers of nearest neighbours can be considered; however, increasing the number of neighbours will decrease the local sensitivity of the method, but it will make the method less sensitive to noise. Instead of $1$-$NN$, the distance to the $k^{th}$ nearest neighbour or the average of the $k$ distances to the first $k$ neighbours can also be used. The value of the threshold (default 1.0) can be changed to either higher or lower values to change the detection sensitivity of the classifier.  \citet{m._j._tax_data_2000} proposes a nearest neighbour method capable of finding data boundaries when the sample size is very low. The boundary thus constructed can be used to detect the targets and outliers. However, this method has the disadvantage that it relies on the individual positions of the objects in the target set. Their method seems to be useful in situations where the data is distributed in subspaces. They test the technique on both real and artificial data and find it to be useful when very small amounts of training data exist (fewer than 5 samples per feature).

\citet{datta_characteristic_1997} modify the standard nearest neighbour algorithm that is appropriate to learn a single class or the positive class. Their modified algorithm, $NNPC$ (nearest neighbour positive class), takes examples from only one class as input. $NNPC$ learns a constant $\delta$ which is the maximum distance a test example can be from any learned example and still be considered a member of the positive class. Any test data object that has a distance greater than $\delta$ from any training data object will not be considered a member of the positive class. The variable $\delta$ is calculated by: 
\begin{equation}
  \delta = Max\{\forall x Min\{\forall y \neq x dist(x,y)\}\}
\end{equation}

where $x$ and $y$ are two examples of the positive class, and Euclidean distance, $dist(x,y)$ is used as the distance function. Datta also experiment with another similar modification called $NNPCN$, that involves learning a vector $<\delta_1,\ldots,\delta_n>$, where $\delta_i$ is the threshold for the $i^{th}$ example. This modification records the distance to the closest example for each example. $\delta_i$ is calculated by:
\begin{equation}
  \delta_i = Min\{\forall y \neq x_i dist(x_i,y)\}
\end{equation}
where $x_i$ is the $i^{th}$ training example. To classify a test example the same classification rule as above is used.

\citet{t._munroe_multi-class_2005} extend the idea of one-class KNN to tackle the recognition of vehicles using a set of features extracted from their frontal view, and present results showing high accuracy in classification. They compare their results with multi-class classification methods and comment that it is not reasonable to draw direct comparisons between the results of the multi-class and single-class classifiers, because the use of training \& testing data sets and the underlying assumptions are quite different. They also note that the performance of multi-class classifier could be made arbitrarily worse by adding those vehicle types to the test set that do not appear in the training set. Since one-class classifiers can represent the concept ``none of the above'', their performance should not deteriorate in these conditions. \citet{g._cabral_novel_2007} propose a one-class nearest neighbour data description using the concept of structural risk minimization. k-Nearest Neighbours (kNN) suffers from the limitation of having to store all training samples as prototypes that would be used to classify an unseen sample. Their paper is based on the idea of removing redundant samples from the training set, thereby obtaining a compact representation aiming at improving generalization performance of the classifier. The results on artificial and UCI datasets show improved performance than the $NN$-$d$ classifiers and also achieved considerable reduction in number of stored prototypes. \citet{g._cabral_combining_2009} present another approach where not only $1$-$NN$ is considered but all of the $k$-nearest neighbours, to arrive at a decision based on majority voting. In their experiments on artificial data, biomedical data\footnote{http://lib.stat.cmu.edu/datasets/ [Accessed: Jan-2012].} and data from the UCI repository, they observe that the $k$-$NN$ version of their classifier outperforms the $1$-$NN$ and is better than $NN$-$d$ algorithms.  \citet{d._gesu_one_2008} present a one-class KNN and test it on synthetic data that simulates microarray data for the identification of nucleosomes and linker regions across DNA. A decision rule is presented to classify an unknown sample X as:

\begin{equation}
 X=
 \begin{cases}
  1, & \text{if} |y \in T_p \quad \text{such that} \quad \delta (y,x) \le \varphi| \ge K\\
  0, & \text{otherwise}
 \end{cases}
\end{equation}

where $T_P$ is the training set for the data object $P$ representing positive instance, $\delta$ is the dissimilarity function between data objects and $j=1$ means that $x$ is positive. The meaning of the above rule is that if there are at least $K$ data objects in $T_P$ with dissimilarity from $x$ no more than $\varphi$, then $x$ is classified as a positive data object, otherwise it is classified as an outlier. This $kNN$ model depends on parameters $K$ and $\varphi$ and their values are chosen by using optimization methods. Their results have shown good recognition rate on synthetic data for nucleosome and linker regions across DNA.

\citet{de_haro-garcia_one-class_2009} use one-class KNN along with other one-class classifiers for identifying plant/pathogen sequences,  and present a comparison of results. They find that these methods are suitable owing to the fact that genomic sequences of plant are easy to obtain in comparison to pathogens, and they build one-class classifiers based only on information from the sequences of the plant. One-class kNN classifiers are used by  \citet{g._glavin_analysis_2009} to study the effect of unexpected outliers that might arise in classification (in their case, they considered the task of classifying spectroscopic data). According to the authors, unexpected outliers can be defined as those outliers that do not come from the same distribution of data as the postive cases or outlier cases in the training data set. Their experiments show that the one-class kNN method is more reliable for the task of detecting such outliers than a similar binary kNN classifier. The `kernel approach' has been used by various researchers to implement different flavours of NN-based classifier for multi-class classification. \citet{s._khan_kernels_2010} extends this idea and propose two variants of one-class nearest neighbour classifiers. These variants use a kernel as a distance metric instead of Euclidean diistance for identification of chlorinated solvents in the absence of non-chlorinated solvents. The first method finds the neighbourhood for nearest $k$ neighbours, while the second method finds $j$ localized neighbourhoods for each of $k$ neighbours. Popular kernels like the polynomial kernel (of degree 1 and 2), RBF and spectroscopic kernels (Spectral Linear Kernel \citep{howley_kernel_2007} and Weighted Spectral Linear Kernel \citep{g._madden._machine_2008}) are used in their experiments. Their kernelized $kNN$ methods perform better than standard $NN$-$d$ and $1$-$NN$ methods and that a one-class classifier with RBF Kernel as distance metric performs better when compared to other kernels.

\paragraph{Bayesian Classifiers.}
\citet{datta_characteristic_1997} suggests a method to learn a Na\"{i}ve Bayes (NB) classifier from samples of positive class data only. Traditional NB attempts to find the probability of a class given an unlabeled data object, $p(C_i|A_1=\nu_1 \& \ldots \& A_n=\nu_n)$. By assuming that the attributes are independent and applying Bayes' theorem the previous calculation is proportional to:
\begin{equation}
  [\prod_{j}^{attributes} p(A_j=\nu | C_i)]p(C_i)
\end{equation}

where $A_j$ is an attribute, $\nu$ is a value of the attribute, $C_i$ is a class, and the probabilities are estimated using the training examples. When only the positive class is available, the calculation of $p(C_i)$ from the above equation cannot be done correctly. Therefore, \citet{datta_characteristic_1997} modifies NB to learn in a single class situation and calls their modification $NBPC$ (Na\"{i}ve Bayes Positive Class) that uses the probabilities of the attribute-values. $NBPC$ computes a threshold $t$ as: 

\begin{equation}
  t=Min[\forall x \prod_{j}^{attributes} p(A_j=\nu_i)]
\end{equation}

where $A_j=\nu_i$ is the attribute value for the example $x$ and $p(A_j=\nu_i)$ is the probability of the attribute's $i^{th}$ value. The probabilities for the different values of attribute $A_j$ is normalized by the probability of the most frequently occurring $\nu$. During classification, if for the test example $\prod_{j}^{attributes} (A_j=\nu_i) \ge t$, then the test example is predicted as a member of the positive class. Datta tests the above positive class algorithms on various datasets taken from UCI repository and conclude that $NNPC$ (discussed in previous section) and $NBPC$ have classification accuracy (both precision and recall values) close to C4.5's value, although C4.5 decision trees are learned from all classes whereas each of the one-class classifiers is learned using only one class. \citet{wang_one_2003} use a simple one-class NB method that uses only positive samples, for masquerade detection in a network. The idea is to generate a user's profile ($u$) using UNIX commands ($c$) and compute the conditional probability $p(c|u)$ for user $u$'s self profile. For the non-self profile, they assume that each command has a random probability $\frac{1}{m}$. For testing a sample $d$, they compare the ratios $p(d|self)$ and $p(d|non\text{-}self)$. The larger the value of this ratio, it is more likely that the command $d$ has come from user $u$.
 
\paragraph{Other Methods.}
\citet{wang_visual_2004} investigate several one-class classification methods in the context of Human-Robot interaction for image classification into faces and non-faces. Some of the important non-standard methods used in their study are Gaussian Data Description, k-Means, Principal Component Analysis, and Linear Programming \citep{pekalska_one-class_2003}, as well as the SVDD. They study the performance of these one-class classification methods on an image recognition dataset and observe that SVDD attains better performance in comparison to the other OCC methods studied, due to its flexibility relative to the other methods, which use very strict models of separation, such as planar shapes. They also investigate the effect of varying the number of features and remark that more features do not always guarantee better results, because with an increase in the number of features, more training data are needed to reliably estimate the class models.  \citet{ercil_one_2002} report a different technique to tackle the OCC problem, based on fitting an implicit polynomial surface to the point cloud of features, to model the target class in order to separate it from the outliers. They show the utility of their method for the problem of defect classification, where there are often plentiful samples for the non-defective class but only very few samples for various defective classes. They use an implicit polynomial fitting technique and show a considerable improvement in the classification rate, in addition to having the advantage of requiring data only from non-defective motors in the learning stage. A fuzzy one-class classifier is proposed by \citet{l._bosco_fuzzy_2009} for identifying patterns of signals embedded in a noisy background. They employed a Multi-Layer Model \citep{d._gesu_multi-layer_2009} as a data processing step and then use their proposed fuzzy one-class classifier for testing on microarray data. Their results show that integrating these two methods could improve the overall classification results. \citet{juszczak_minimum_2009} propose a one-class classifier that is built on the minimum spanning tree of only the target class. The method is based on a graph representation of the target class to capture the underlying structure of the data. This method performs distance-based classification as it computes the distance from a test object to its closest edge.  They show that their method performs well when the data size is small and has high dimensions. \citet{gayar_weighted_2010} suggest that the performance of the method of Juszczak et al. is reduced in the presence of outliers in the target class. To circumvent this problem, they present two bagging ensemble approaches that are aimed to reduce the influence of outliers in the target training data and show improved performance on both real and artificially contaminated data. \citet{hempstalk_one-class_2008} present a one-class classification method that combines a density estimator to draw a reference distribution and a standard model for estimating class probability for the target class. They use the reference distribution to generate artificial data for the second class and decompose this problem as a standard two-class learning problem and show that their results on UCI dataset and typist dataset are comparable with standard OSVM \citep{scholkopf_support_2000}, with the advantage of not having to specify a target rejection rate at the time of training. \citet{silva_hypergraph-based_2009} present a high dimension anomaly detection method that uses limited amount of unlabeled data. Their algorithm uses a variational Expectation-Maximization (EM) method on the hypergraph domains that allows edges to connect more than two vertices simultaneously. The resulting estimate can be used to compute degree of anomalousness based on a false-positive rate. The proposed algorithm is linear in the number of training data objects and does not require parameter tuning. They compare the method with OSVM and another $K$-point entropic graph method on synthetic data and the very high dimensional Enron email database, and conclude that their method outperforms both of the other methods.

\subsection{Category 3: Application Domain Applied}
\subsubsection{Text / Document Classification} 
\label{text}
Traditional text classification techniques require an appropriate distribution of positive and negative examples to build a classifier; thus they are not suitable for the problem of OCC. It is of course possible to manually label some negative examples, though depending on the application domain, this may be a labour-intensive and time-consuming task. However, the core problem remains, that it is difficult or impossible to compile a set of negative samples that provides a comprehensive characterization of everything that is `not' the target concept, as is assumed by a conventional binary classifier. It is a common practice to build text classifiers using positive and unlabeled examples\footnote{Readers are advised to refer to survey paper by \citet{zhang_learning_2008}} (in \textit{semi-supervised} setting\footnote{Readers are advised to refer to survey paper on semi-supervised learning by \citet{zhu_semi_05}}) as collecting unlabeled samples is relatively easy and fast in many text or Web page domains \citep{nigam_text_2000, liu_partially_2002}. In this section, we will discuss some of the algorithms that exploit this methodology with application to text classification.

The ability to build classifiers without negative training data is useful in a scenario if one needs to extract positive documents from many text collections or sources. \citet{nigam_text_2000} show that accuracy of text classifiers can be improved by adding small amount of labelled training data to a large available pool of unlabelled data. They introduce an algorithm based on EM and NB. The central idea of their algorithm is to train the NB classifier based on the labelled documents and then probabilistically label the unlabeled documents and repeat this process till it converges. To improve the performance of the algorithm, they propose two variants that give weights to modulate the amount of unlabeled data, and use mixture components per class. They show that using this type of methodology can result in reduction of error by factor of up to 30\%. \citet{liu_building_2003} study the problem of learning from positive and unlabeled data and suggest that many algorithms that build text classifiers are based on two steps: 
\begin{enumerate}
  \item  Identifying a set of reliable/strong negative documents from the unlabeled set. In this step, Spy-EM \citep{liu_partially_2002} uses a Spy technique, PEBL \citep{yu_pebl:_2004} uses a technique called 1-DNF \citep{yu_pebl:_2002}, and Roc-SVM \citep{liu_building_2003} uses the Rocchio algorithm \citep{rocchio_relevant_1971}.
  
  \item Building a set of classifiers by iteratively applying a classification algorithm and then selecting a good classifier from the set. In this step, Spy-EM uses the Expectation Maximization (EM) algorithm with an NB classifier, while PEBL and Roc-SVM use SVM. Both Spy-EM and Roc-SVM have same methods for selecting the final classifier. PEBL simply uses the last classifier at convergence, which can be a poor choice.
\end{enumerate}

These two steps together work in an iterative manner to increase the number of unlabeled examples that are classified as negative, while at the same time maintain the correct classification of positive examples. It was shown theoretically by \citet{yu_pebl:_2002} that if the sample size is large enough, maximizing the number of unlabeled examples classified as negative while constraining the positive examples to be correctly classified will give a good classifier.  \citet{liu_building_2003} introduce two new methods, one for Step 1 (i.e. the NB) and one for Step 2 (i.e. SVM alone) and perform an evaluation of all 16 possible combinations of methods for Step 1 and Step 2 that were discussed above. They develop a benchmarking system called LPU (Learning from Positive and Unlabeled Data)\footnote{http://www.cs.uic.edu/~liub/LPU/LPU-download.html [Accessed: Jan-2012]} and propose an approach based on a biased formulation of SVM that allows noise (or error) in positive examples. Their experiments on Reuters and Usenet articles suggest that the biased-SVM approach outperforms all existing two-step techniques. 

\citet{yu_text_2003} explore SVMC \citep{yu_svmc:_2003} (for detail on this technique refer to Section \ref{osvm}) for performing text classification without labelled negative data. They use Reuters and WebKb\footnote{http://www.cs.cmu.edu/afs/cs.cmu.edu/project/theo-20/www/data/ [Accessed: Jan-2012]} corpora for text classification and compare their method against six other methods: (i) Simple Mapping Convergence (MC); (ii) OSVM; (iii) Standard SVM trained with positive examples and unlabeled documents substituted for negative documents; (iv) Spy-EM; (v) NB with Negative Noise; and (vi) Ideal SVM trained from completely labelled documents. They conclude that with a reasonable number of positive documents, the MC algorithm gives the best performance among all the methods they considered. Their analysis show that when the positive training data is not under-sampled, SVMC significantly outperforms other methods because SVMC tries to exploit the natural gap between positive and negative documents in the feature space, which eventually helps to improve the generalization performance.  \citet{peng_text_2006} present a text classifier from positive and unlabeled documents based on Genetic Algorithms (GA) by adopting a two stage strategy (as discussed above). Firstly, reliable negative documents are identified by an improved 1-DNF algorithm. Secondly, a set of classifiers are built by iteratively applying the SVM algorithm on training data objects sets. They then discuss an approach to evaluate the weighted vote of all classifiers generated in the iteration steps, to construct the final classifier based on a GA. They comment that the GA evolving process can discover the best combination of the weights. Their experiments are performed on the Reuters data set and compared against PEBL and OSVM and it is shown that the GA based classification performs better.

\begin{figure}
  \centering
    \includegraphics[width=3.5in]{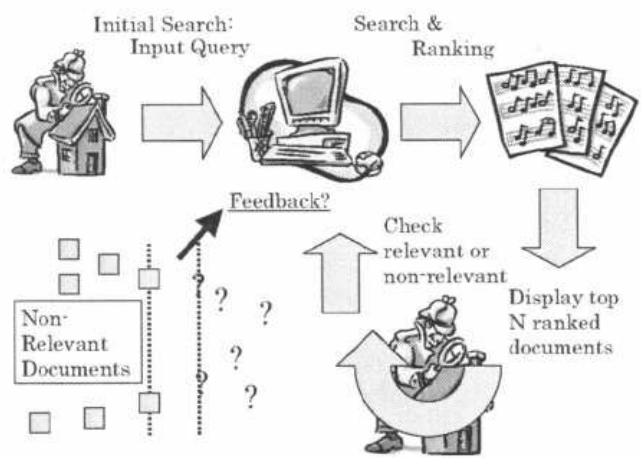}
  \caption{Outline of a problem in the relevance feedback documents retrieval. Source: \citet{onoda_one_2005}.}
  \label{fig:onoda}
  \end{figure}

\citet{koppel_authorship_2004} study the \textit{Authorship Verification} problem where only examples of writings of a single author is given and the task is to determine if given piece of text \textit{is} or \textit{is not} written by this author. Traditional approaches to text classification cannot be applied directly to this kind of classification problem. Hence, they present a new technique called `unmasking' in which features that are most useful for distinguishing between books A and B are iteratively removed and the speed with which cross-validation accuracy degrades is gauged as more features are removed. The main hypothesis is that if books A and B are written by the same author, then no matter what differences there may be between them (of genres, themes etc), the overall essence or regularity in writing style can be captured by only a relatively small number of features. For testing the algorithm, they consider a collection of twenty-one 19th century English books written by 10 different authors and spanning a variety of genres and obtain overall accuracy of 95.7\% with errors almost equally distributed between false positives and false negatives.  \citet{onoda_one_2005} report a document retrieval method using non-relevant documents. Users rarely provide a precise query vector to retrieve desired documents in the first iteration. In subsequent iterations, the user evaluates whether the retrieved documents are relevant or not, and correspondingly the query vector is modified in order to reduce the difference between the query vector and documents evaluated as relevant by the user. This method is called relevance feedback. The relevance feedback needs a set of relevant and non-relevant documents to work usefully. However, sometimes the initial retrieved documents that are presented to a user do not include relevant documents. In such a scenario, traditional approaches for relevance feedback document retrieval systems do not work well, because the system needs relevant and non relevant documents to construct a binary classifier (see Figure \ref{fig:onoda}). To solve this problem, Onoda et al. propose a feedback method using information from non-relevant documents only, called non-relevance feedback document retrieval. The design of non-relevance feedback document retrieval is based on OSVM \citep{scholkopf_estimating_1999}. Their proposed method selects documents that are discriminated as not non-relevant and that are near the discriminant hyper-plane between non-relevant document and relevant documents. They compare the proposed approach with conventional relevance feedback methods and vector space model without feedback and show that it consistently gives better performance as compared to other methods. \citet{pan_one-class_2008} extend the concept of classifying positive examples with unlabeled samples in the Collaborative Filtering (CF) application. In CF, the positive data is gathered based on user interaction with the web like news items recommendation or bookmarking pages, etc. However, due to ambiguous interpretations, limited knowledge or lack of interest of users, the collection of valid negative data may be hampered. Sometime negative and unlabeled positive data are severely mixed up and it becomes difficult to discern them.  Manually labelling negative data is not only intractable considering the size of the web but also will be poorly sampled. Traditional CF algorithms either label negative data, or assume missing data are negative. Both of these approaches have an inherent problem of being expensive and biased to the recommendation results. \citet{pan_one-class_2008} propose two approaches to one-class CF to handle the negative sparse data to balance the extent to which to treat missing values as negative examples. Their first approach is based on weighted low rank approximation \citep{srebro_weighted_2003} that works on the idea of providing different weights to error terms of both positive and negative examples in the objective function. Their second approach is based on sampling missing values as negative examples. They perform experiments on real world data from social bookmarking site del.icio.us and Yahoo News data set and show that their method outperforms other state of the art CF algorithms. 

\citet{denis_text_2002} introduce a NB algorithm and shows its feasibility for learning from positive and unlabeled documents. The key step in their method is the estimation of word probabilities for the negative class, because negative examples were not available. This limitation can be overcome by assuming an estimate of the positive class probability (the ratio of positive documents in the set of all documents). In practical situations, the positive class probability can be empirically estimated or provided by domain knowledge. Their results on WebKB data set  show that error rates of NB classifiers obtained from $p$ positive data objects, $PNB$ (Positive Na\"{i}ve Bayes), trained with enough unlabeled examples are lower than error rates of NB classifiers obtained from $p$ labeled documents. \citet{denis_text_2003} consider situations where only a small set of positive data is available together with unlabeled data. Constructing an accurate classifier in these situations may fail because of the shortage of properly sampled data. However, learning in this scenario may still be possible using the co-training framework \citep{blum_combining_1998} that looks for two feature views over the data. They propose a Positive Na\"{i}ve Co-Training algorithm, $PNCT$, that takes a small pool of positive documents as its seed. $PNCT$ first incrementally builds NB classifiers from positive and unlabeled documents over each of the two views by using PNB. Along the co-training steps, self-labelled positive examples and self-labelled negative examples are added to the training sets. A base algorithm is also proposed which is a variant of $PNB$ that is able to use these self-labelled examples. The experiments on the WebKB dataset show that co-training algorithms lead to significant improvement of classifiers, even when the initial seed is only composed of positive documents. \citet{calvo_learning_2007} extend the $PNB$ classification idea to build more complex Bayesian classifiers in the absence of negative samples when only positive and unlabeled data are present. A positive tree augmented NB ($PTAN$) in a positive-unlabeled scenario is proposed along with the use of a Beta distribution to model the apriori probability of the positive class, and it is applied to $PNB$ and $PTAN$. The experiments suggest that when the predicting attributes are not conditionally independent, $PTAN$ performs better than $PNB$. The proposed Bayesian approach to estimating apriori probability of the positive class also improves the performance of $PNB$ and $PTAN$.  \citet{he_naive_2010} improves the $PNB$ by selecting the value of the prior probability of the positive class on the validation set using a performance measure that can be estimated from positive and unlabeled examples. Their experiments suggest that the proposed algorithm performs well even without the user specifying the prior probability of the positive class.

\citet{pan_nearest_2010} propose two variants of a nearest neighbour classifier for classification of uncertain data under learning from positive and unlabeled scenario. The method outperforms the $NN$-$d$ \citep{m._j._tax_data_2000} and OCC method by \citet{hempstalk_one-class_2008}.  \citet{elkan_learning_2008} show that if a classifier is trained using labelled and unlabeled data then its predicted probabilities differ from true probabilities only by a constant factor. They test their method on theapplication of  identifying protein records and show it performs better in comparison to the standard biased SVM method \citep{liu_building_2003}. \citet{zhang_one-class_2008} propose a one-class classification method for the classification of text streams with concept drift. In situations where text streams withe large volumes of documents arrive at high speed, it is difficult to label all of them and few positive samples are labelled. They propose a stacked ensemble approach and compare it against other window-based approaches and demonstrate its better performance. \citet{blanchard_semi-supervised_2010} present a different outlook on learning with positive and unlabeled data that develops general solution to this problem by a surrogate problem related to Neyman-Pearson classification, which is a binary classification problem subject to a constraint on false-positive rate while minimizing the false-negative rate. It is to be noted that in their problem formulation, outliers are not assumed to be rare but that special case is also discussed. They perform theoretical analysis to deduce generalization error bounds, consistency, and rates of convergence for novelty detection and show that this approach optimally adapts to unknown novelty distributions, whereas the traditional methods assume a fixed uniform distribution. Their experiments compare the proposed method with OSVM on several datasets\footnote{http://www.csie.ntu.edu.tw/~cjlin/libsvmtools/datasets/  [Accessed: July-2013]} and find comparable performance. Learning from positive and unlabeled data has been studied in other domains as well apart from text classification such as facial expression recognition \citep{cohen_learning_2003}, gene regulation networks \citep{cerulo_learning_2010} etc.

\subsubsection{Other Application Domains}
In this subsection we will highlight some of the other applications of one-class classification methods that may not necessarily employ learning from positive and unlabeled data. Some of these areas are: Handwriting Detection \citep{m._j._tax_uniform_2001, m._j._tax_one-class_2001, scholkopf_support_2000, hempstalk_one-class_2008}; Information Retrieval \citep{m._manevitz_one-class_2001}; Missing Data/Data Correction \citep{juszczak_combining_2004, xiaomu_svm-based_2008}; Image Database Retrieval \citep{m._j._tax_uniform_2001, chen_one-class_2001, gondra_improving_2004,  seo_application_2007}; Face/Object Recognition Applications \citep{wang_visual_2004, zeng_one-class_2006, bicego_face_2005}, Remote Sensing \citep{li_positive_2011}; Stream Mining \citep{zhang_one-class_2008, zhang_learning_2010, liu_one-class_2011}; Chemometrics and Spectroscopy \citep{kittiwachana_one_2010, g._glavin_analysis_2009, s._khan_kernels_2010, xu_automated_2007, hao_identification_2010}; Biometrics \citep{bergamini_combining_2009, bergamini_fusion_2008}; Assistive Technologies \citep{yang_wearable_2010, khan_towards_2012}; Time Series Analysis \citep{sachs_one-class_2006, n._nguyen_positive_2011}; Disease Detection \citep{cohen_application_2004, zhang_combining_2011}; Medical Analysis \citep{gardner_one-class_2006, zhou_extraction_2005}; Bioinformatics \citep{j._spinosa_svms_2004, a._c._p._l._ferreira_de_carvalho_combining_2005, t._alashwal_one-class_2006, wang_psol:_2006, yousef_learning_2008}; Steganalysis \citep{lyu_steganalysis_2004, m._rodriguez_steganography_2007}; Spam Detection \citep{sun_novel_2005, t._wu_using_2005, m._schneider_learning_2004}; Audio Surveillance, Sound \& Speaker Classification \citep{brew_evaluation_2007, rabaoui_improved_2007, rabaoui_using_2008}; Ship Detection \citep{tang_one-class_2005};  Vehicle Recognition \citep{t._munroe_multi-class_2005}; Collision Detection \citep{m._quinlan_application_2003}; Anomaly Detection 
\citep{li_improving_2003, a._tran_one-class_2004, zhang_one_2007,perdisci_using_2006, yilmazel_leveraging_2005, v._nguyen_application_2002}; Intrusion Detection \citep{giacinto_network_2005,f._evangelista_fuzzy_2005, luo_research_2007}; Credit Scoring \citep{kennedy_credit_2009}; Yeast Regulation Prediction \citep{kowalczyk_one_2002}; Customer Churn Detection \citep{zhao_customer_2005}; Relevant Sentence Extraction \citep{kruengkrai_using_2003}; Machine Vibration Analysis \citep{m_._j._tax_support_1999}; Machine Fault Detection \citep{ercil_one_2002, m._j._tax_support_2004, j._shin_one-class_2005, sarmiento_fault_2005}; and Recommendation Tasks \citep{yasutoshi_one-class_2006}. Compression neural networks for one-class classification have been used to detect mineral deposits \citep{skabar_single-class_2003} and for fMRI Analysis \citep{r._hardoon_one-class_2005, r._hardoon_fmri_2005}. One-class Fuzzy ART networks have been explored to classify cancerous cells \citep{murshed_classification_1996}.   

\section{Conclusions and Open Research Questions}
\label{conclusions}
The goal of OCC algorithms is to induce generalized classifiers when only one class (the target or positive class) is well characterized by the training data and the negative or outlier class is either absent, poorly sampled or the negative concept is not well defined. The limited availability of data makes the problem of OCC more challenging and interesting. The research in the field of OCC encompasses several research themes developed over time. In this paper, we have presented a unified view on the general problem of OCC and presented a taxonomy for the study of OCC problems. We have observed that the research carried out in OCC can be broadly represented by three different categories or areas of study, which depends upon the availability of training data, classification algorithms used and the application domain investigated. Based on the categories under the proposed taxonomy, we have presented a comprehensive literature survey of current state-of-the-art and significant research work in the field of OCC,  discussing the techniques used and methodologies employed with a focus on their limitations, importance and applications.

Over the course of several years, new OCC algorithms have emerged and new application areas have been exploited. Although the OCC field is becoming mature,  there are still several fundamental problems that are open for research, not only in describing and training classifiers, but also in scaling, controlling errors, handling outliers, using non-representative sets of negative examples, combining classifiers, generating sub-spaces, reducing dimensionality and making a fair comparison of errors with multi-class classification.

In the context of OCC, we believe that classifier ensemble methods need further exploration. Although there exist several bagging models, new techniques based on boosting and random subspace warrant further attention. Random subspace methods with one-class variants of decision trees and nearest neighbour classifiers can be an interesting research direction. The Random oracle Ensemble \citep{i._kuncheva_classifier_2007}  has been shown to fare better than standard ensembles for the multi-class classification problems; however, for OCC it has been not explored. We believe that carrying out research on new ensemble methods within the domain of OCC can bring interesting results.

Another point to note here is that in OSVMs, the kernels that have been used mostly are Linear, Polynomial, Gaussian or Sigmoidal. We suggest it would be fruitful to investigate some more innovative forms of kernels, for example Genetic Kernels \citep{howley_evolutionary_2006} or domain specific kernels, such as Weighted Linear Spectral Kernel \citep{howley_kernel_2007}, that have shown greater potential in standard SVM classification. Moreover, parameter tuning of standard kernels may give biased results, therefore we believe that researchers should focus on efficiently tuning and optimizing kernel parameters. The kernels used as distance metric have shown promising results in the basic form of one-class nearest neighbour classifiers. We believe further research in this direction can show interesting insights into the problem. An important issue that has been largely ignored by OCC researchers is how to handle missing data in the target class and still be able to develop robust one-class classifiers. This kind of scenario makes the OCC problem more difficult. Khan et al. \citep{khan_bayesian_2012} address the issue of handling missing data in target class by using Bayesian Multiple Imputations \citep{b._rubin_multiple_1987} and EM and propose several variants of one-class classifier ensemble that can perform better than traditional method of mean imputation. However, we recommend that research involving these and other advanced data imputation methods can help in building one-classifiers that can handle missingness in the data. Feature selection for one-class classifiers is a difficult problem because it is challenging to model the behaviour of features for one class in terms of their discriminatory power. There are some studies in this direction \citep{d._villalba_evaluation_2007}; however, a lack of advanced methods and techniques to handle high dimensional positive class data can still be a bottleneck in learning one-class classifiers.

In the case where abundant unlabeled examples and some positive examples are available, researchers have used many different two-step algorithms, as have been discussed in Section \ref{text}. We believe that a Bayesian Network approach to such OCC problems would be an interesting research area. When learning on only positive data using NB method, an important problem that hampers the flexibility of the model is the estimation of prior probabilities. An important research direction is to put in effort to estimate prior distribution with minimal user intervention. A possible direction in this attempt is to explore mixtures of beta distribution for inferring prior class probability \citep{calvo_positive_2008}.

Normally OCC is employed to identify anomalies, outliers, unusual or unknown behaviours, and failing to identify such novel observations may be costly in terms of risks associated with health, safety and money. In general, most of the researchers assume equal cost of errors (false alarms and miss alarms), which may not be true in the case of OCC. However, traditional cost-sensitive learning methods \citep{ling_cost_2007} may not fit here because neither the prior probability of the outlier class nor the associated cost of errors are known. Data-dependent approaches that deduce cost of errors from the training data objects may not generalize the cost across different domains of even same application area. In the papers we reviewed, only one research paper discusses the cost-sensitive aspect of OCC \citep{luo_research_2007} and it shows that this area of research is largely unexplored. We believe that approaches based on careful application of Preference Elicitation techniques \citep{chen_survey_2004} can be useful to deduce cost of errors. In terms of data types, most of the research work in OCC is focused on numerical or continuous data, however not much emphasis is given to categorical or mixed data approaches. Similarly,  adaptation and development of OCC methods for streaming data analysis and online classification also need more research effort. Many of the OCC algorithms perform density estimation and assume that outliers are uniformly distributed in low density regions, however if the target data also lies in low density region than such methods may start rejecting positive data objects or if the thresholds are increased than will start accepting outliers. To tackle such scenarios, alternate formulations of OCC is required.

In terms of applications of OCC, we have seen that it has been explored in many diverse fields with very encouraging results. With the emergence of Assistive Technologies to help people with medical conditions, OCC has a potential application to map patients' individual behaviour under different medical conditions which would otherwise be very difficult to handle with multi-class classification approaches. Activity recognition is also a central problem in such domains. The data for these domains is normally captured by sensors that are prone to noise and may miss vital recordings. Multi-class classifiers are difficult to employ in these applications to detect unknown or anomalous behaviours, such as those related to the user/patient or the sensor itself. We believe OCC can play an important role in modelling these kinds of applications. In activity recognition and some computer vision problems, even the collected positive data may contain instances from negative class (e.g. extracting text and non-text regions from PowerPoint slides). Building binary classes on such unclean data is detrimental to the classification accuracy. We believe that employing OCC classifiers as a post-processing to filter out the noise from target data objects can be conducive to build better multi-class classifiers. Finally, we strongly recommend the development of open-source OCC software, tools and benchmark datasets that can be used by researchers to compare and validate results in coherent and systematic way.

This survey provides a unified and in-depth insight into the current study in the field of OCC. Depending upon the data availability, algorithm use and applications domain, appropriate OCC techniques can be applied and improved upon. We hope that the proposed taxonomy along with this survey will provide researchers with a direction to formulate future novel work in this field. 

\paragraph{Acknowledgements}
The authors are grateful to Dr. D.M.J. Tax, Dr. L. Manevitz, Dr. K. Li, Dr. H. Yu and Dr. T. Onoda for their kind permission to reproduce figures from their respective papers.

\bibliography{references}
\bibliographystyle{plainnat}
\end{document}